%% file: neurips_2024.tex
\pdfoutput=1
\documentclass{article}

\usepackage[preprint, nonatbib]{neurips_2024}

\usepackage[utf8]{inputenc} % allow utf-8 input
\usepackage[T1]{fontenc}    % use 8-bit T1 fonts
\usepackage{hyperref}       % hyperlinks
\usepackage{url}            % simple URL typesetting
\usepackage{booktabs}       % professional-quality tables
\usepackage{amsfonts}       % blackboard math symbols
\usepackage{nicefrac}       % compact symbols for 1/2, etc.
\usepackage{microtype}      % microtypography
\usepackage{xcolor}         % colors
\usepackage{graphicx}
\usepackage{amsmath}
\usepackage{wrapfig}
\usepackage{caption}
\usepackage{subcaption}
\usepackage[numbers]{natbib}
\usepackage{listings}

\input{math_commands}

\input{latex_def}

\title{Towards Data-Centric RLHF: Simple Metrics for Preference Dataset Comparison}

\author{%
  Judy Hanwen Shen\thanks{Work done during internship at Apple} \\
  Stanford University\\
  \texttt{jhshen@stanford.edu} \\
  \And
  Archit Sharma  \\
  Stanford University \\
  \texttt{architsh@stanford.edu} \\
  \AND
  Jun Qin \\
  Apple \\
  \texttt{jqin22@apple.com} \\
}

\begin{document}

\maketitle

\begin{abstract}
\input{abstract.tex}
\end{abstract}

\input{main_text.tex}

\bibliography{references}
\bibliographystyle{abbrvnat}
\newpage
\appendix
\input{supplementary.tex}

\end{document}

%% file: math_commands.tex
\usepackage{amsmath,amsfonts,bm, bbm}

\def\eqref#1{equation~\ref{#1}}
\def\1{\bm{1}}

\DeclareMathAlphabet{\mathsfit}{\encodingdefault}{\sfdefault}{m}{sl}
\SetMathAlphabet{\mathsfit}{bold}{\encodingdefault}{\sfdefault}{bx}{n}

\DeclareMathOperator*{\argmax}{arg\,max}

%% file: latex_def.tex
\newcommand{\Er}[2][]{\mathbb{E}_{#1}\left[#2\right]}
\newcommand{\ignore}[1]{}

%% file: abstract.tex
The goal of aligning language models to human preferences requires data that reveal these preferences. Ideally, time and money can be spent carefully collecting and tailoring bespoke preference data to each downstream application. However, in practice, a select few publicly available preference datasets are often used to train reward models for reinforcement learning from human feedback (RLHF). While new preference datasets are being introduced with increasing frequency, there are currently no existing efforts to measure and compare these datasets. In this paper, we systematically study preference datasets through three perspectives: scale, label noise, and information content. We propose specific metrics for each of these perspectives and uncover different axes of comparison for a better understanding of preference datasets. Our work is a first step towards a data-centric approach to alignment by providing perspectives that aid in training efficiency and iterative data collection for RLHF.

%% file: main_text.tex
\section{Introduction}
Reinforcement learning from human feedback (RLHF) is typically the final stage of the modern large language model (LLM) training pipeline~\cite{achiam2023gpt, touvron2023llama,groeneveld2024olmo}. The reward models necessary for RLHF algorithms are predominantly trained from datasets of pairwise preferences \cite{bai2022training, ouyang2022training}. While a substantial number of works have focused on new algorithms for learning from preference data to better train reward models~\cite{moskovitz2023confronting, zheng2023secrets, dong2024rlhf, xiong2024iterative}, relatively few works have examined qualities of these datasets themselves. At the very minimum, all of these pairwise datasets of human preferences contain examples with 1) a prompt, 2) two responses, and 3) an annotation of which response is preferred. Beyond this basic structure, preference datasets vary widely in domain (e.g. code, chat, QA, etc.), generation process (e.g. synthetic vs human),  collection procedure (e.g. annotation, prompt generation), and even size (e.g. 10k - 300k examples~\cite{zheng2023secrets, cui2023ultrafeedback}). 

Ideally, a custom preference dataset for each specific application can be collected, and carefully labeled by multiple annotators for reward model training. New technical reports that accompany state-of-the-art language models highlight the importance of preference data quality yet give little to no details about the preference datasets used~\cite{2405.04434, 2407.21783}. Among publicly available preference datasets, there is folk wisdom that more carefully curated datasets are better, yet no rigorous study or methodology for comparing these datasets exists beyond summary statistics, such as token count~\cite{dong2024rlhf}. Today, little is known about when and why one preference dataset may be better than another, nor what ``better'' can mean in the context of these datasets. 

In this paper, we initiate the study of measuring properties of preference datasets for the purpose of reward model training. A useful measurement should be robust to different base model choices and applicable to any dataset containing pairwise preferences. To this end, we propose three data-centric approaches for comparing preference datasets: effective sample size, noise invariance, and information content. We evaluate both in-distribution performance and domain generalization (i.e. through a standard reward modeling benchmark) on the induced reward model trained on these datasets. We validate our results through ablations across different model sizes to demonstrate the connection between these measurements and subsequent reward model performance. Together, our work gives three simple but intuitive perspectives for understanding preference datasets that are broadly applicable to the development of new datasets across domains. 

\section{Related Work}
\paragraph{Data-Centric Methods}
Scaling laws introduced to describe the relationship between parameters, data, and compute for pre-training have been widely accepted as the explanation for why larger models and more data are better for language model training~\cite{kaplan2020scaling, hoffmann2022training}. Different approaches for improving data quality and composition have been proposed as efficient alternatives for indiscriminately training on all available data~\cite{penedo2024fineweb, xie2024doremi}. However, the scale of pre-training data vastly eclipses the scale of data used in the fine-tuning and RLHF stages. Data quality and data selection for reward model training may be more similar to supervised learning settings than language modeling. In supervised learning and supervised fine-tuning, careful data selection and pruning have been shown to lower the number of samples required~\cite{paul2021deep, sorscher2022beyond, xia2024less}. However, reward models do differ from the supervised learning setting since they are adapted from these pre-trained base models. Recent work has studied data scaling for fine-tuning LLMs to find that LLM performance benefits more from pre-training data scaling than fine-tuning data scaling and the optimal fine-tuning method is task and data-dependent~\cite{zhang2024scaling}.  

\paragraph{Publicly Available Preference Datasets}
% early preference datasets that accompany the suggestion of rlhf
For RLHF preference datasets in particular, early works collected datasets on the order of tens of thousands of examples for reward model training. For example, for a summarization task Stienon et al.,~\cite{stienon2020learning} collected 64k preference pairs based on Reddit prompts, while the WebGPT~\cite{nakano2021webgpt} reward model was trained with 16k preference pairs based on prompts from existing QA datasets. Subsequent datasets follow a more general human-assistant format while being much larger (e.g. OpenAssistant~\cite{kopf2024openassistant}, HH-RLHF~\cite{bai2022training}, Stanford Human Preferences~\cite{pmlr-v162-ethayarajh22a}). However, these datasets vary drastically in collection procedure. For example, for InstructGPT and HH-RLHF humans were asked to rank model-generated responses while for OpenAssistant and Stanford Human Preferences preferences for different human-generated responses were gathered. More recently, preference datasets where both responses and rankings are synthetically generated have gained popularity~\cite{cui2023ultrafeedback, daniele2023amplify-instruct}. These synthetically constructed datasets offers more training samples and more diversity in terms of the topics generated. There is also a movement back to creating smaller but carefully annotated preferences, often with multiple annotators~\cite{wang2024helpsteer2}. Despite the large variation in practices for generating these different datasets, there has been little comparison and characterization of how different datasets affect reward model training. 

\paragraph{Challenges of Reward Modeling and Learning from Human Preferences}
Defining data quality is complex for preference data since many different tasks may use the same reward model for RLHF. There are concerns with the representativeness of preferences as well as the alignment between collected data and the intended objective~\cite{lambert2023alignment, kirk2024prism, chen2024preference}. One suggestion for measuring the effectiveness of reward models is standardized benchmarks on reward model performance on a variety of common tasks~\cite{lambert2024rewardbench}. This approach measures the generalization of a single reward model on different tasks by testing how well each reward model performs on scoring the chosen response higher. The top-performing models on this benchmark leaderboard include models of a variety of sizes from 8B to 340B parameters and a variety of preference data sizes from $10$k to more than $700$k examples. Given this mishmash of different approaches, it is important to understand how to measure preference data quality for the reward modeling step of RLHF. This work aims to characterize the elements of preference data quality that inform practical decisions around data generation, annotation, and usage in this setting.

% One Story: 
\section{Model Agnostic Data Metrics}
\subsection{Preliminaries}
Let $x$ be the prompt, $y_w$ be the winning (chosen) response, and $y_l$ be the losing (rejected) response. Let $D = \{(x, y_w, y_l)_i\}_{i=1}^n \sim \mathcal{D}$ be the dataset of preferences that we will study. Let $r: \mathcal{X} \times \mathcal{Y} \rightarrow \mathbb{R}$ be the reward model that maps a $(x, y)$  prompt response pair to a score. In reward modeling, we want to compare the rewards of two given generations. The Bradley-Terry model defines $Y_{ij}$ as a Bernoulli random variable representing the outcome of whether the completion $y_i$ is preferred or wins over the completion $y_j$. Under this model, $Y_{ij} \sim$ Bernoulli($p_{ij}$) and the log ratio of the probability that $y_i$ wins over $y_j$ is:
\[
\log \frac{p_{ij}}{1 - p_{ij}} = r(x, y_i) - r(x, y_j). 
\]
If we let $y_i$ be the winning completion $y_w$ and $y_j$ be the losing completion $y_l$, we can then write the probability of the reward model preferring $y_w$ as: 
\[
P(y_w \succ y_l) = \frac{\exp(r(x, y_w))}{\exp(r(x, y_l)) + \exp(r(x, y_w))}.
\]
Following prior work~\cite{ouyang2022training, bai2022training}, the probability of the reward model giving a higher score to the chosen response can then be maximized directly through the following objective function: 
\[
L = -\Er[(x, y_w, y_l) \sim \mathcal{D}]{\log \sigma(r(x, y_w) - r(x, y_l))}.
\]

\subsection{Datasets and Models}
We examine four publicly available preference datasets in our study: Anthropic Helpful-Harmless (\textsc{HH-RLHF})~\cite{bai2022training}, Ultrafeedback (\textsc{UltraFeedback})~\cite{cui2023ultrafeedback}, LMSYS Arena Preferences (\textsc{Lmsys})~\cite{chiang2024chatbot}, and PKU-SafeRLHF (\textsc{SafeRLHF})~\cite{ji2024pku}. These datasets are selected based on their frequent use in prior works \cite{bai2022training, dong2024rlhf}\footnote{We include dataset details in the supplementary materials}. For each dataset, we examine their behavior on reward models trained from pre-trained models of different sizes: 350 million (Opt-350m~\cite{zhang2022opt}), 1 billion (TinyLlama-1B-3T~\cite{zhang2024tinyllama}), and 7 billion parameters (Llama2-7B and Llama2-7B-Chat~\cite{touvron2023llama}). We focus predominately on reward models trained from base models but also include ablations with fine-tuned versions since the practice around reward model training varies. For example, some papers train reward models from checkpoints already fine-tuned with instructions and human feedback (e.g. Llama3-8B-Instruct)~\cite{dong2024rlhf}) and other works train reward models directly from based models~\cite{wang2024helpsteer2, zheng2023secrets}. Notably, Ouyang et. al. \cite{ouyang2022training} remark that similar reward model quality was observed between training on a base 6B model and an instruction-tuned 6B model. We evaluate both in-domain and generalization performance through evaluation set accuracy and Rewardbench~\cite{lambert2024rewardbench} respectively.

\section{Experiments}

\subsection{Scaling: Are larger preference datasets better?}
\begin{figure}
    \centering
    \includegraphics[width=\linewidth,height=5cm,keepaspectratio]{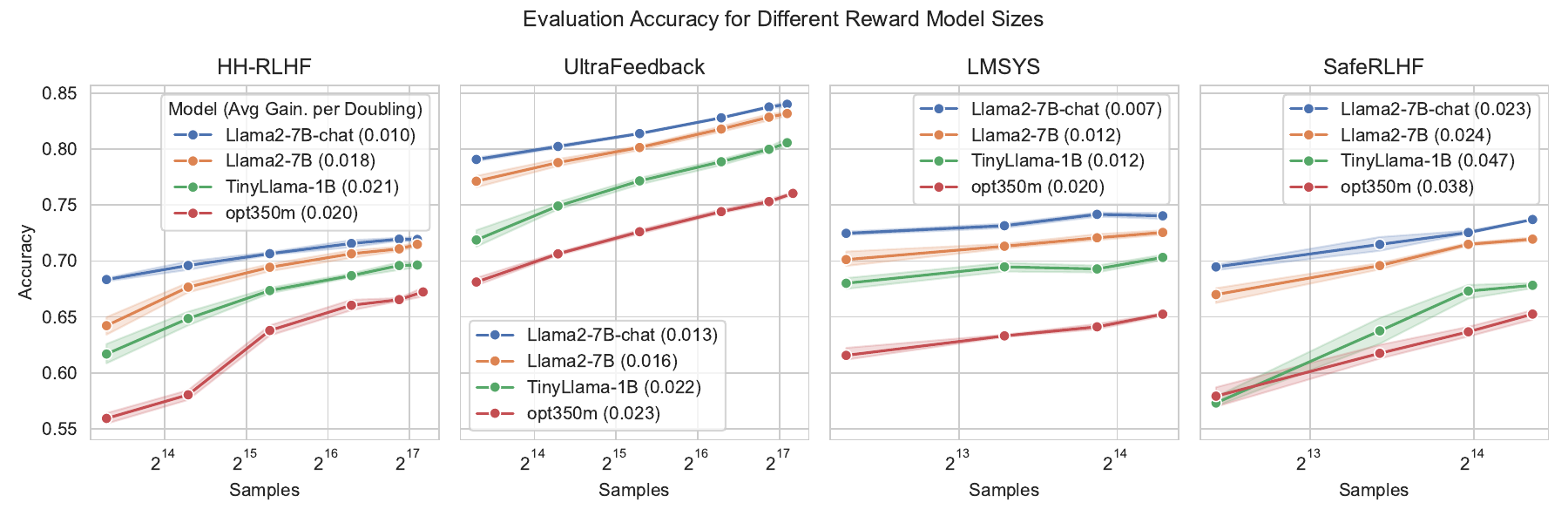}
    \caption{Scaling behavior when measuring evaluation set accuracy is dataset dependent.}
    \label{fig:in-domain-scaling}
\end{figure}

The first perspective we examine is the role of dataset size for different preference datasets. Unlike scaling laws for pre-training, there is no consensus about how large a preference dataset should be to train a good reward model. For summarization in particular, Stienon et. al.~\cite{stienon2020learning} estimate that doubling their particular dataset size leads to a 1.1\% increase in reward model validation accuracy until 65k examples. In contrast, others have found that even when using 2.9 million examples, reward model accuracy continues to improve~\cite{touvron2023llama}. While these differences can be blamed on the dataset composition, the impact of increasing the training set size across different datasets has not been studied. We examine four datasets that range in size from 30k examples to 200k examples and observe how training dataset size impacts performance. Figure \ref{fig:in-domain-scaling} illustrates the impact of scaling on evaluation set accuracy. For all datasets, the larger models (Llama2-7B, Llama2-7B-chat), gain less from doubling the dataset size. While Llama2-7B-chat is fine-tuned with RLHF from part of \textsc{HH-RLHF}, this pattern remains even for other datasets that were released after Llama2-7B-chat. Among datasets, \textsc{SafeRLHF} has the highest average gain per doubling of the training dataset (2.4-4.7\%) for all models. 

\begin{figure}
    \centering
    \includegraphics[width=\linewidth]{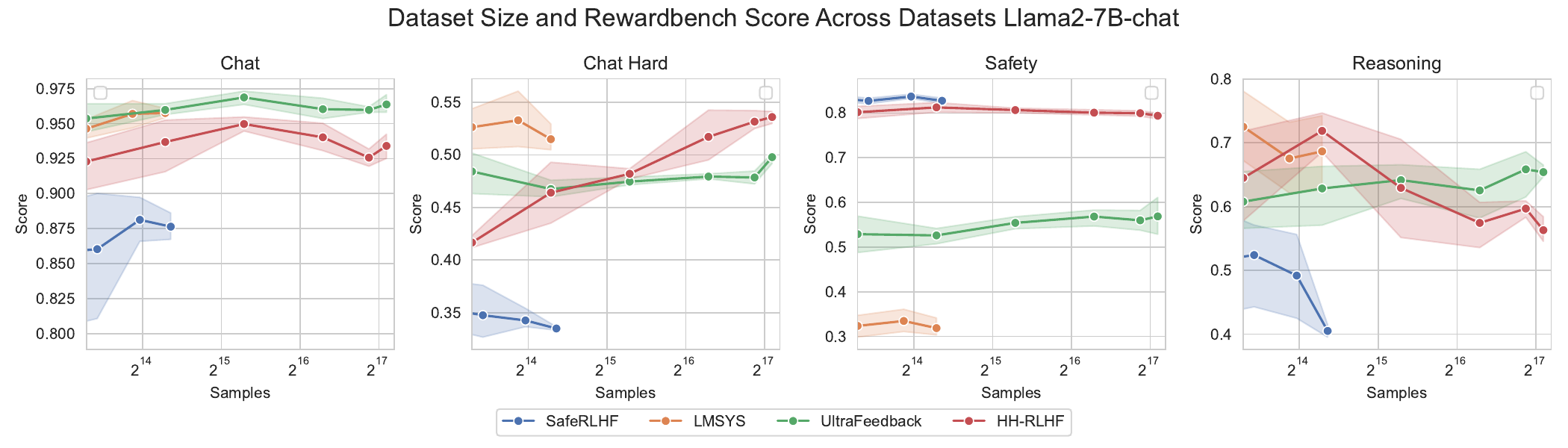}
    \caption{Comparing RewardBench performance across different datasets for Llama2-7B-chat model. Increasing the dataset size does not improve performance for most datasets on most tasks.}
    \label{fig:ood-scaling}
\end{figure}

We also investigate the effect of increasing dataset size on a more general suite of reward model tasks that might be outside the training distribution using RewardBench~\cite{lambert2024rewardbench} (Figure \ref{fig:ood-scaling}). Unlike evaluation accuracy, increasing dataset size does not always improve, and sometimes harms, performance on this benchmark. Some datasets dominate a task across all sample sizes (e.g. \textsc{UltraFeedback} on Chat, \textsc{HH-RLHF} on Reasoning, and \textsc{SafeRLHF} on Safety). This shows that a small subset of samples (e.g. 10K examples or 10\% of a dataset) is already sufficient and that dataset composition may be more important in achieving good performance than scale. For example, 10k examples from \textsc{SafeRLHF} outperforms 140k examples from \textsc{HH-RLHF}. These results are model invariant across different reward model sizes. For example, \textsc{UltraFeedback} remains the best dataset for the Chat category across both the 350M and 1B model\footnote{See Appendix~\ref{app:scaling} for details}. 

\subsection{Noise Invariance: How robust are reward models to label noise?}
Prior works have reported the human agreement with the collected preferences to be $~76\%$ for summarization~\cite{stienon2020learning} and $~73\%$ inter-annotator agreement for response quality for general instruction tasks~\cite{ouyang2022training}. Ideally, annotator disagreement serves as a filter for low-quality preference data, however, even if the collection process is unknown, it is still useful to understand how much noise there might be in the preference dataset. In image classification tasks, neural networks are robust to label noise~\cite{rolnick2017deep}. In these settings, a random label is used instead of the true label in multiclass classification. In the context of preference data, we can model label noise as the flipping of the chosen response with the rejected response. We can define: $p$ as the noise rate and add random label noise by constructing a dataset: 
\begin{align*}
    (x, y_w, y_l) = \begin{cases}
    (x, y_l, y_w) & \text{w.p.} \quad p \\
    (x, y_w, y_l) & \text{w.p.} \quad 1-p.
\end{cases}
\end{align*}
%If the dataset is noisier, we should see lower noise invariance since flipping labels fewer labels is sufficient to critically affect the performance. 
Table~\ref{tab:noise_table} shows the percentage of the peak evaluation set accuracy achieved when 30\% of labels are flipped. Overall, we find that reward model performance remains unaffected by label flipping until 30-40\% of labels are flipped. The same pattern is observed on RewardBench tasks across all models\footnote{Full plots and more details can be found in Appendix \ref{app:noise}}. 
\begin{table}[]
    \centering
    \small
    \begin{tabular}{l|llll}
    \hline
    Base Model   & \textsc{HH-RLHF} & \textsc{UltraFeedback} & \textsc{LMSYS} & \textsc{SafeRLHF} \\     \hline
    Opt-350m        & 88.6\%  & 95.0\%   & 94.9\% & 92.6\%   \\
    TinyLlama-1B    & 90.1\%  & 95.4\%   & 95.4\%  & 94.4\%  \\
    Llama2-7B       & 78.9\%  & 93.3\%   & 94.6\% & 84.9\% \\  
    Llama2-7B-chat  & 92.7\%  & 93.6\%   & 92.4\% & 90.7\%  
    \end{tabular}
    \caption{\small Percentage of total evaluation accuracy achieved with 30\% of labels flipped for each dataset for different sized reward models. \textsc{UltraFeedback} and \textsc{LMSYS} are particularly noise invariant but all datasets are fairly robust to label flipping.}
    \label{tab:noise_table}
\end{table}

\begin{wrapfigure}{r}{0.5\textwidth}
  \begin{center}
    \includegraphics[width=0.4\textwidth]{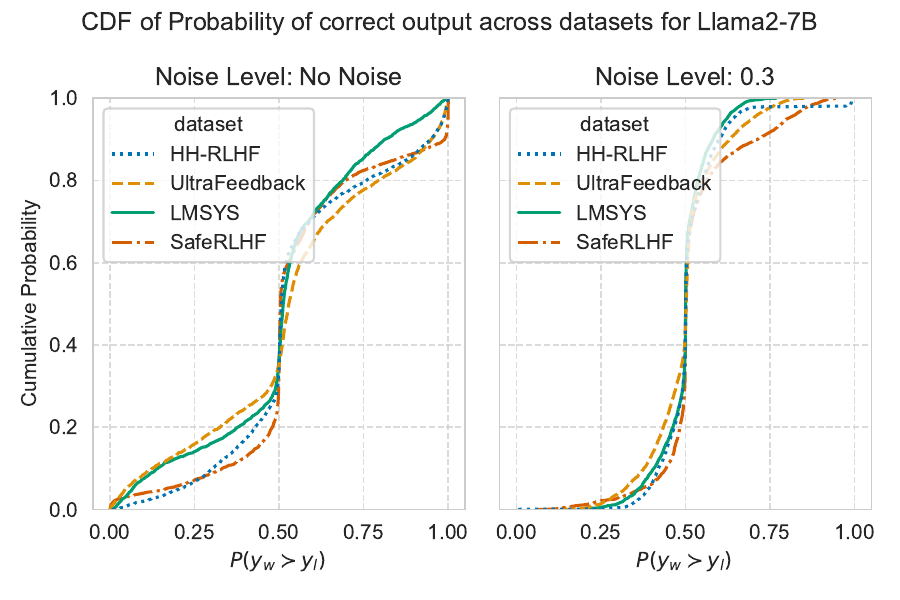}
  \end{center}
  \caption{\small Empirical CDF of $P(y_w \succ y_l)$ for different datasets at different noise levels for Llama 7B on RewardBench. When there is no noise, some datasets induce a more confident distribution even with the same number of training examples. As more noise is added, all probabilities shift towards 0.5 and the datasets become indistinguishable}
  \label{fig:noise-dataset}
\end{wrapfigure}

\begin{figure}
    \centering
    \includegraphics[width=\linewidth]{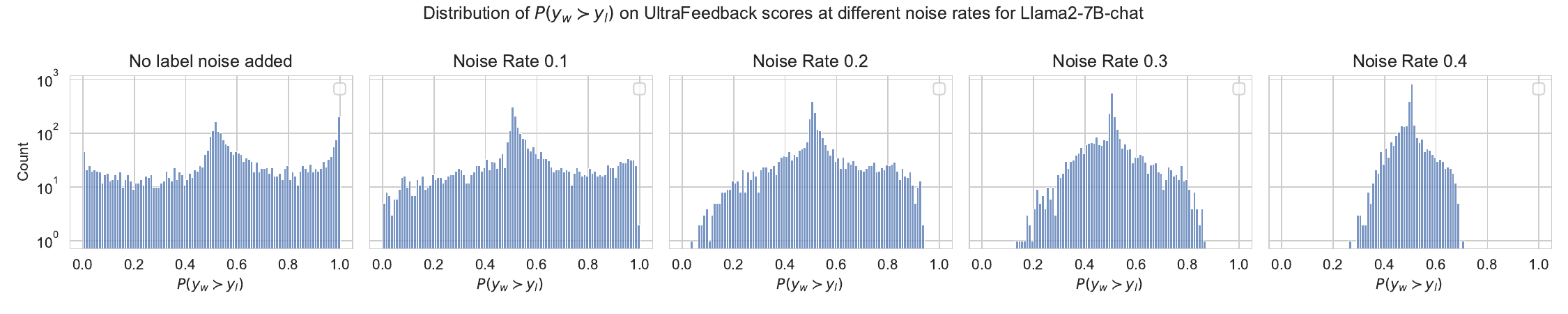}
    \caption{\small The impact of noise on reward model confidence $P(y_w \succ y_l)$ on \textsc{UltraFeedback} for RewardBench. We see that as the noise rate (\% of flipped labels) increases, the probability of the winning response being chosen concentrates around $0.5$. This phenomenon is similar across all models and datasets to different extents.}
    \label{fig:noise-prob}
\end{figure}

\paragraph{Explaining Noise Invariance: The Role of Noise in Reward Model Confidence}
We can look at the underlying prediction probabilities to further understand why introducing label noise does not significantly affect performance both on the evaluation set and the RewardBench tasks. Since accuracy for both sets of metrics is calculated through expected binary outcomes (i.e. $\Er[(x, y_w, y_l) \sim \mathcal{D}]{\mathbbm{1}[\hat{y} = y_c]}$ where $\hat{y} = \argmax_{y \in \{y_w, y_l\}}{r(x, y)}$), we can use the Bradley-Terry model to calculate $P(y_w \succ y_l)$ and investigate how these distributions change. As the noise rate increases, the distribution of probabilities (e.g. $P(y_w \succ y_l)$) becomes more concentrated around 0.5 (Figure \ref{fig:noise-prob}). This pattern is consistent across different reward model sizes and datasets. 
Across different datasets, Figure~\ref{fig:noise-dataset} shows that when label noise is introduced, \textsc{HH-RLHF} and \textsc{LMSYS} collapses quicker to $P(y_w \succ y_l) \approx 0.5$ than other datasets. This suggests that there might be a higher level of baseline noise in the \textsc{HH-RLHF} labels that results in more uncertain predictions. This pattern is again consistent across different reward model sizes. 

To precisely characterize model confidence, we can measure the expected calibration error (ECE) of reward model outputs~\cite{guo2017calibration}. However, in the Bradly-Terry model, using $P(y_w \succ y_l)$ directly as model confidence results in perfect accuracy when $P(y_w \succ y_l)>0.5$. The only prior work we could find that measures calibration in reward models uses $\max\{P(y_w \succ y_l), P(y_l \succ y_w)\}$ as the confidence of the model~\cite{pikus2023baseline}. To properly measure calibration, we can write each evaluation pair as $(x, y_1, y_2, z)$ and split it into $(x, y_w, y_l, z=1)$ and $(x, y_l, y_w, z=0)$. Then to calculate the calibration error we can use $P(z=1):=P(y_1 \succ y_2)$ as model confidence and plot the count of $z=1$ as the accuracy (see Figure ~\ref{fig:reliability-diagram-llama7B}). The overall ECE is equivalent to the max method from prior work but now we have confidence values in the entire interval of $[0,1]$ instead of just $[ 0.5, 1]$. As label noise increases, we observe lower calibration error (e.g. ECE=0.183 no noise to ECE=0.086 30\% label noise for \textsc{UltraFeedback}) (see Section ~\ref{sec:calibration} for more details).    
\begin{figure}
     \centering
     \begin{subfigure}[b]{0.48\textwidth}
         \centering
         \includegraphics[width=\textwidth]{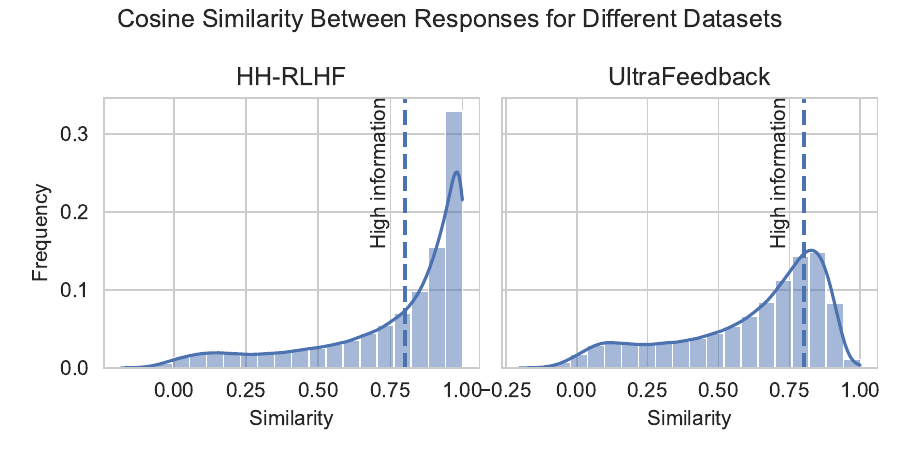}
         \label{fig:similarity-cosine}
     \end{subfigure}
     \hfill
     \begin{subfigure}[b]{0.48\textwidth}
         \centering
         \includegraphics[width=\textwidth]{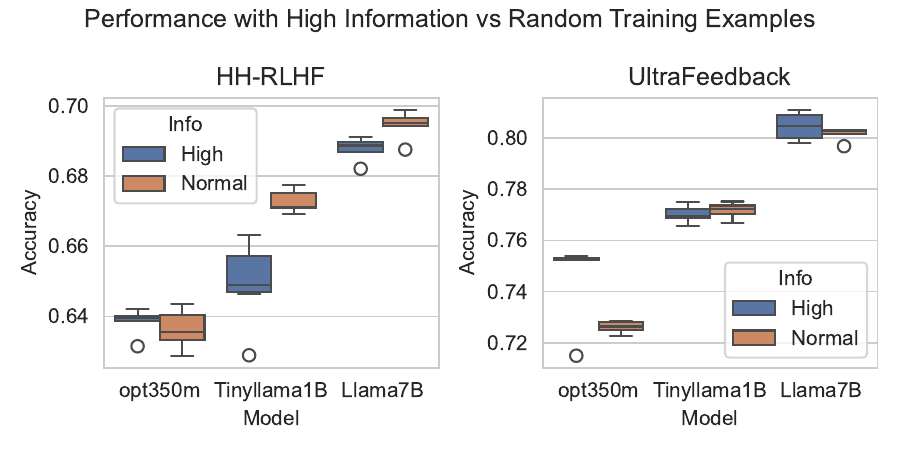}
         \label{fig:high-info}
     \end{subfigure}
    \caption{\small (Left) Distribution of cosine similarity of response pairs for different datasets. The \textsc{HH-RLHF} dataset contains much more similar response pairs (e.g. $(y_w, y_l)$) than the \textsc{UltraFeedback} dataset. (Right) The evaluation set accuracy for training different models with ``high information'' or low response similarity data compared to a random sample. The benefits of ``high information'' are most salient in the smallest model.}
    \label{fig:info-exp}
\end{figure}

\subsection{Information Content: Are high contrast responses necessary for reward model learning?}
A major dichotomy in how preference datasets are generated is whether the responses are human-written or sampled from large language models. For example, the Anthropic Helpful-Harmless (\textsc{HH-RLHF}) dataset contains response pairs generated from responses from LLMs of the same family~\cite{bai2022training}. In contrast, the Stanford Human Preference Dataset (\textsc{SHP}) dataset is gathered from pairs of (presumably human) Reddit responses~\cite{pmlr-v162-ethayarajh22a}. As responses are more similar in quality, prior work has found that human annotation agreement reduces these responses~\cite{touvron2023llama}. While the relative informativeness of an example for training a reward model is likely model-dependent, since the models used for reward model training vary in training data, a minimal level of contrast between the chosen and rejected response is likely a prerequisite for valuable examples in preference datasets. Given the differences in response generation, we can compare and contrast different datasets by computing the cosine similarity between embeddings of responses (i.e. $1-d_{cos}(y_w, y_l)$)\footnote{We use all-MiniLM-L6-v2 from Sentence Transformer to generate embeddings. We investigated a suite of different sentence embeddings and found them to be highly correlated.}. Figure~\ref{fig:info-exp} shows that the \textsc{HH-RLHF} dataset has many more similar response pairs than \textsc{UltraFeedback}. To understand the impact of training with high-information examples, we created a threshold of 0.8 in cosine similarity and designated the examples with a smaller similar as ``high information". Fixing the training set size, we compared the performance of training the high-information examples to a random sample. Surprisingly, the results vary by model and dataset. For the larger models (i.e. 1B+ parameters), there is little difference between the high information and random training sets of the same size. However, for the smaller 350 million parameter model, we see that the high information examples often resulted in a better evaluation accuracy (Figure \ref{fig:info-exp}).

\section{Discussion}
Our work investigates three aspects of preference datasets to identify dataset differences and connect these differences to downstream performance on both in- and out-domain tasks. Firstly, we find that while preference datasets vary in size, a larger dataset is not better than a smaller dataset that is more relevant to the task. Furthermore, increasing dataset size gives only marginal gains for in-domain evaluation accuracy and may even hinder performance on out-of-domain tasks. Future work introducing new preference datasets should report the marginal gain of using the entire dataset on different models compared to using just 10-25\% of the dataset. 

Secondly, we find all four of the preference datasets we examine to be extremely noise invariant. We attribute this observation to label noise introducing more uncertainty in reward model predictions rather than prediction reversal. This suggests that better preference datasets can tolerate a higher level of label noise. Future work introducing new preference datasets should report the noise invariance of a dataset and the calibration error induced in the downstream reward model. 

Lastly, we find that preference datasets vary widely in the distribution of similarity of response pairs. The performance improvements of training from high information or dissimilar response pairs depends on the underlying reward model. An extreme case is if the underlying language model has undergone RLHF policy learning using a preference dataset, then the relative value or information of this dataset should be lower for reward modeling. Recent work has proposed that learning policies from on-policy data outperforms methods using out-of-distribution data ~\cite{tajwar2024preference}. Future work should define and investigate on-policy data for reward model learning in the context of RLHF.

%% file: supplementary.tex
\section{Noise: Calibration and Reward Modeling}
\label{sec:calibration}
\begin{figure}
    \centering
    \includegraphics[width=\linewidth]{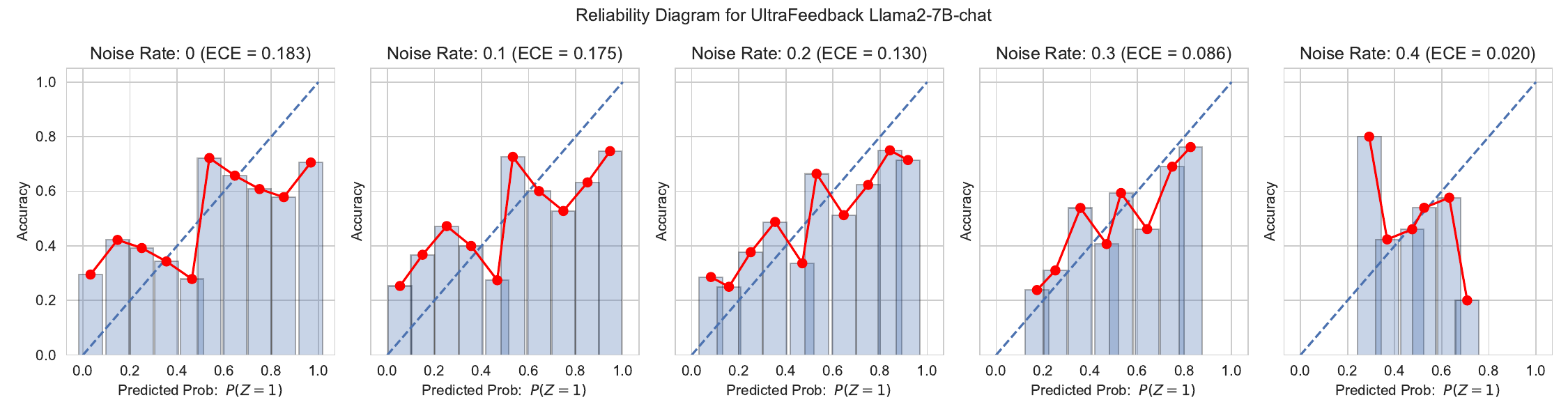}
    \caption{Reliability diagram illustrating expected calibration error (ECE) at different levels of noise for \textsc{UltreaFeedback} on RewardBench examples. More noise decreases calibration error.}
    \label{fig:reliability-diagram-llama7B}
\end{figure}

Thus far, very few works have explored the notion of calibration in reward models. For pairwise preferences, we can think of a reward model as a binary predictor where the notion of calibration is rather natural. Expected calibration error, while suffering from real drawbacks~\cite{blasiok2023unifying}, is the most commonly used metric for measuring miscalibration~\cite{guo2017calibration}. To compute calibration error, bins can created such that for a bin $B_m$, the confidence of the bin is just averaged the predicted probability: 

\[
\text{conf}(B_m) = \frac{1}{B_m} \sum_{i\in B_m} \hat{p}_i,
\]
and the accuracy of bin $B_m$ is the average accuracy of samples in the confidence bin range: 
\[
\text{acc}(B_m) = \frac{1}{B_m} \sum_{i\in B_m} 1[\hat{y}_i = y_i],
\]
where $y_i$ is the predicted label. In this setup, a perfectly calibrated predictor would have matching confidence and accuracy for each bin. In other words, the expected calibration error is the difference between the accuracy and confidence in each bin: 
\[
ECE = \sum_{m=1}^M\frac{|B_m|}{n} | \text{acc}(B_m) - \text{conf}(B_m) |
\]
A problem arises when computing this quantity for reward models on preference data if $\hat{p}_i$ is naively taken to be $\hat{p}_i = P(y_{w_i} \succ y_{l_i})$. This is because by definition, if $P(y_{w_i} \succ y_{l_i}) >0.5$, $\hat{y}_i = y_i$. This means that zero calibration error can only be achieved through $P(y_{w_i} \succ y_{l_i})\in \{0, 1\}$ with a perfect predictor. 

The only prior work that studies calibration in reward models suggests computing the model probability as~\cite{pikus2023baseline}: 
\[\hat{p}_i = \max\{P(y_{w_i} \succ y_{l_i}), P(y_{l_i} \succ y_{w_i})\}.\] 
This gives the right intuition that if $\hat{p}_i \approx 0.5$: we should be very uncertain of the outcome. However, this approach restricts $\hat{p}_i \in [0.5, 1]$. Thus, we suggest an alternative approach in creating another random variable $z \in \{0, 1\}$ to randomize the label outcomes so that each example has the following format $(x, y_1, y_2, z)$. Each example, $(x, y_w, y_l)$ becomes the following two examples: $(x, y_w, y_l,z=1)$ and $(x, y_l, y_w,z=0)$. Now we have the confidence of a bin as: 
\[
\text{conf}(B_m) = \frac{1}{B_m} \sum_{i\in B_m} Pr[z_i = 1] = Pr(y_1 \succ y_2)
\]
and the accuracy of a bin as: 
\[
\text{acc}(B_m) = \frac{1}{B_m} \sum_{i\in B_m} z_i
\]
This approach gives us the reliability diagram in Figure \ref{fig:reliability-diagram-llama7B}. We can see that as label noise increases, calibration error decreases. A trivial predictor can achieve zero ECE by always predicting the average of the labels. The figure shows that as a dataset approaches 50\% noise, $Pr[z_i = 1]$ collapses to values near 0.5 for all examples. We encourage future work to continue investigating the calibration of reward models through our proposed method. 

This transformation we suggest can be done in five simple lines of code:

% \begin{lstlisting}[language=Python]
%     p_chosen = sigmoid(w_rewards - l_rewards)   #P(yw > yl)
%     chosen_labels = np.ones(len(p_chosen)) 
%     p_rejected = 1-p_chosen
%     rejected_labels = 1-chosen_labels
%     # Some function that computes the ece given probabilities and true labels
%     compute_ece(y_pred=np.concatenate([p_chosen, p_rejected]),
%     y_true=np.concatenate([chosen_labels, rejected_labels])) 
% \end{lstlisting}

\section{Scale: Percentile Saturation}
\begin{figure}
    \centering
    \includegraphics[width=\linewidth]{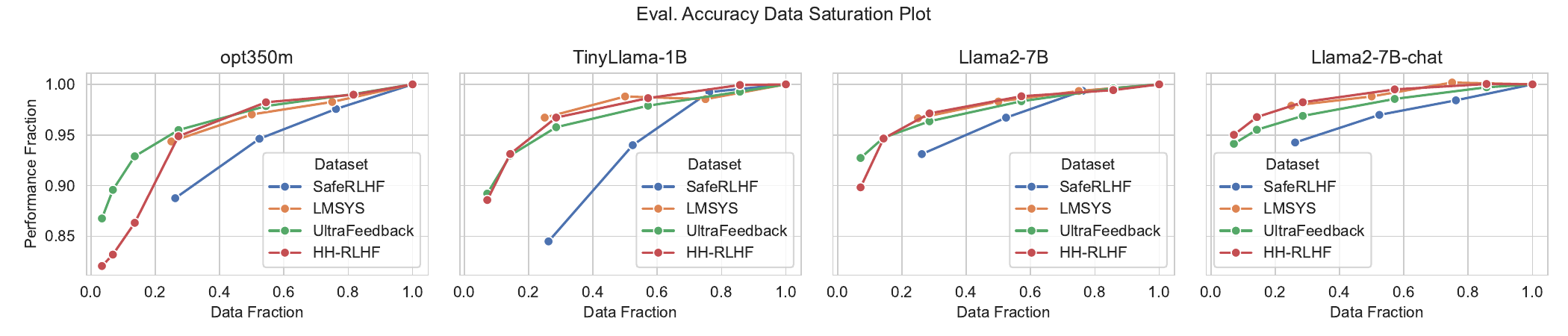}
    \caption{Comparing evaluation accuracy data fraction vs performance fraction, \textsc{SafeRLHF} is the slowest to achieve $>95\%$ of total accuracy, requiring at least 50\% of the dataset. In comparison, other datasets like \textsc{HH-RLHF} only require 10-25\% of the dataset depending on the model.}
    \label{fig:sat}
\end{figure}

Our work compares preference datasets of vastly different sizes. In our main paper, we present two approaches, the \textit{scaling law approach} of looking at how the performance of each dataset changes with increasing data (Figure~\ref{fig:in-domain-scaling}) and the \textit{benchmarking approach} where we plot the performance of different datasets for each task (Figure~\ref{fig:ood-scaling}). Here we would like to present a third choice of \textit{data saturation curves}. On the y-axis we plot the percentage of total performance achieved and on the x-axis we plot the percentage of total data used. This allows us to compare the data efficiency of datasets. In Figure~\ref{fig:sat}, the first observation we can make is that while the shape of the slope of each line becomes flatter with large models, the ordering of datasets remains the same. This allows us to observe that across models of vastly different sizes, \textsc{SafeRLHF} is a dataset that is not very redundant. This is not an artifact of dataset size since \textsc{LMSYS} is approximately the same size. 

%TODO: 
%\section{Noise and Information Content}

\section{Dataset and Experiment Details}
Our work looks at 4 different openly available preference datasets. We excluded preference datasets collected or derived from Reddit data due to recent restrictions with respect to terms of service. Specifically, the four datasets we used came from the following hugging face dataset URLs: 
\begin{itemize}
    \item \textsc{HH-RLHF}: \url{Anthropic/hh-rlhf} 
    \item \textsc{UltraFeedback}: \url{RLHFlow/UltraFeedback-preference-standard}
    \item \textsc{LMSYS} \url{lmsys/lmsys-arena-human-preference-55k}
    \item \textsc{SafeRLHF} \url{RLHFlow/PKU-SafeRLHF-30K-standard}
\end{itemize}
To ensure minimal data discrepancies between models, we filtered out examples longer than 512 tokens according to each model tokenizer. We also removed ties from the \textsc{LMSYS} dataset. 

\subsection{Response Pair Distances}
\begin{figure}
    \centering
    \includegraphics[width=\linewidth]{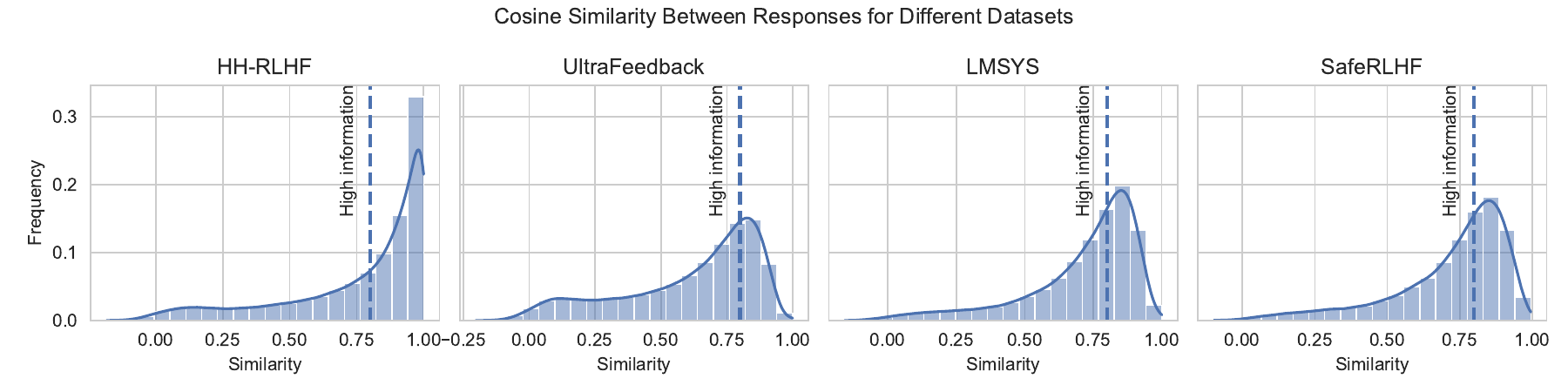}
    \caption{Distribution of Cosine Similarity of winning and losing responses across datasets. \textsc{HH-RLHF} contains many more similar pairs than other datasets}. 
    \label{fig:emb_distance_all}
\end{figure}

For computing distances between responses, we compared several different sentence embeddings. We compared instruction embeddings~\cite{su2022one}, retrieval embeddings\cite{nussbaum2024nomic}, as well as general-purpose embeddings\cite{lan2019albert, reimers-2019-sentence-bert}. We found that cosine and Euclidian distances derived from all of them were highly correlated. Thus, we used a general-purpose pre-trained model: \texttt{ all-MiniLM-L6-v2}. Using embeddings from this model, Figure~\ref{fig:emb_distance_all} shows the contrast in response similarity between different datasets. We see that \textsc{HH-RLHF} contains many more similar winning-losing response pairs compared to other datasets. Furthermore, even though \textsc{LMSYS} responses are generated from a much more diverse set of models than the other three datasets, there are still more similar responses than dissimilar responses. We expect forum-based preference datasets such as Stanford Human Preferences to follow a vastly different distribution of response similarity. 

\begin{figure}
    \centering
    \includegraphics[width=\linewidth]{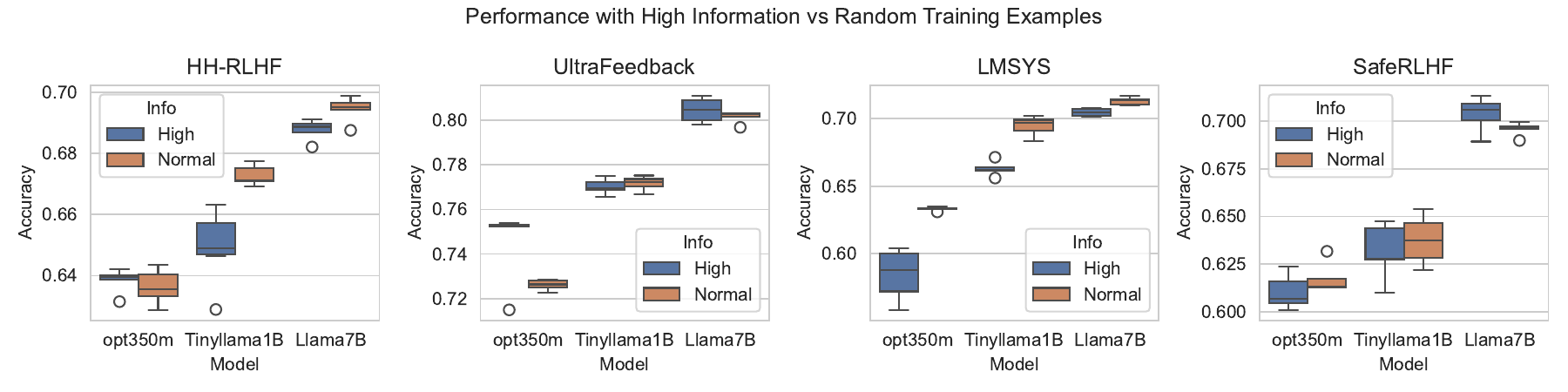}
    \caption{The evaluation set accuracy for training different models with “high information” or low response similarity data
compared to a random sample. The benefits of “high information” are most salient in the smallest model.}
    \label{fig:high-info-all}
\end{figure}

Designating a threshold of 0.8 in similarity, we consider examples that are below 0.8 to be high information. Training on a subset of high-information examples, we compare the downstream performance with a random sample of the training set. While our initial hypothesis may be that training with high information examples would benefit downstream performance, we see that this is only true for small models such as opt350m (Figure~\ref{fig:high-info-all}). One explanation for this effect is that the embeddings used are trained with only 1B pairs on a $\le$33M\footnote{\url{https://huggingface.co/microsoft/MiniLM-L12-H384-uncased}} parameter model. Once reward models are adapted from base modes with billions of parameters trained with trillions of tokens, these metrics of similarity might not be useful. An alternative explanation is that the value or information content of pairs of examples may depend on the base model itself. Future work should investigate model-dependent data valuation for preference data.

\section{Complete Results}
\label{app:results}
\subsection{Dataset Scaling}
\label{app:scaling}
\begin{figure}
    \centering
    \includegraphics[width=\linewidth]{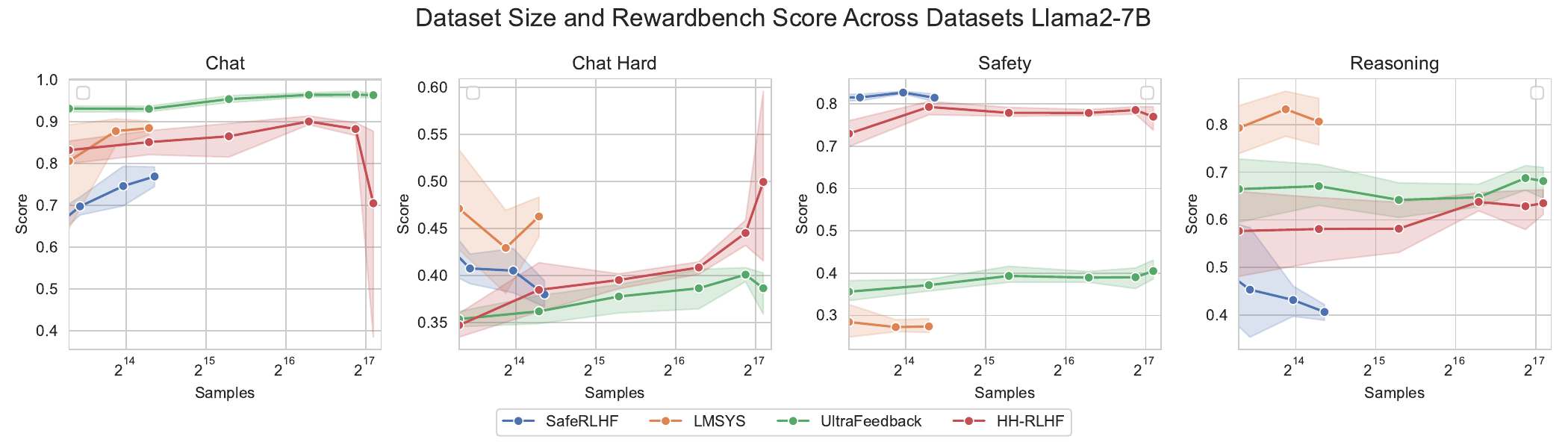}
    \caption{Comparing RewardBench performance across different datasets for Llama2-7B Model. Increasing the dataset size is insufficient to close the performance gap between datasets the best dataset depends on the evaluation task within RewardBench.}
    \label{fig:ood-scaling-7B}
\end{figure}

\begin{figure}
    \centering
    \includegraphics[width=\linewidth]{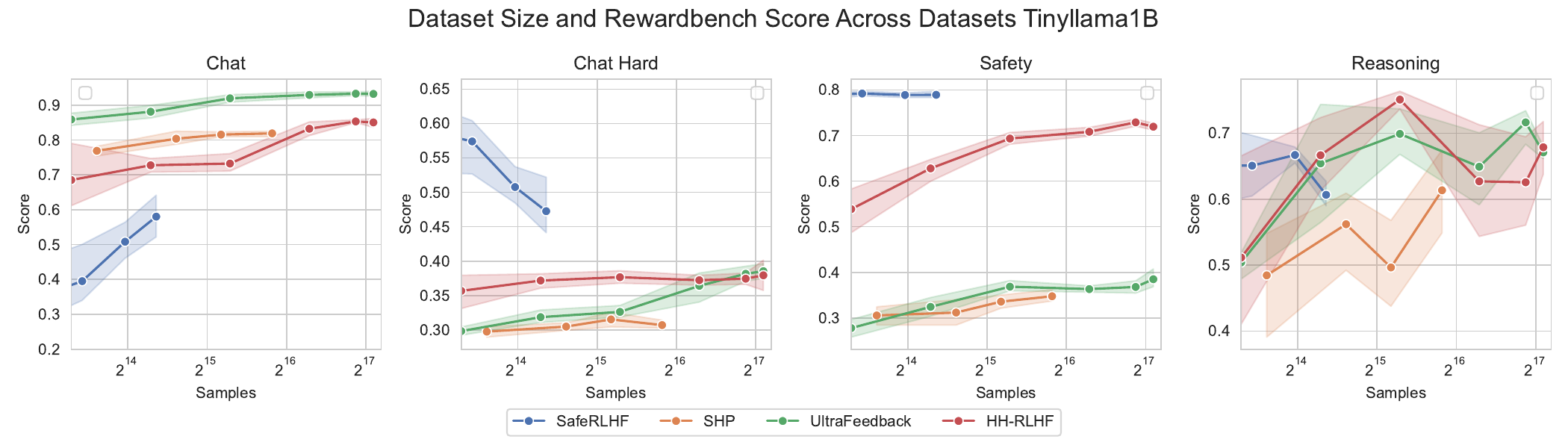}
    \caption{Comparing RewardBench performance across different datasets for TinyLlama-1B Model. Increasing the dataset size is insufficient to close the performance gap between datasets the best dataset depends on the evaluation task within RewardBench.}
    \label{fig:ood-scaling-1B}
\end{figure}

\begin{figure}
    \centering
    \includegraphics[width=\linewidth]{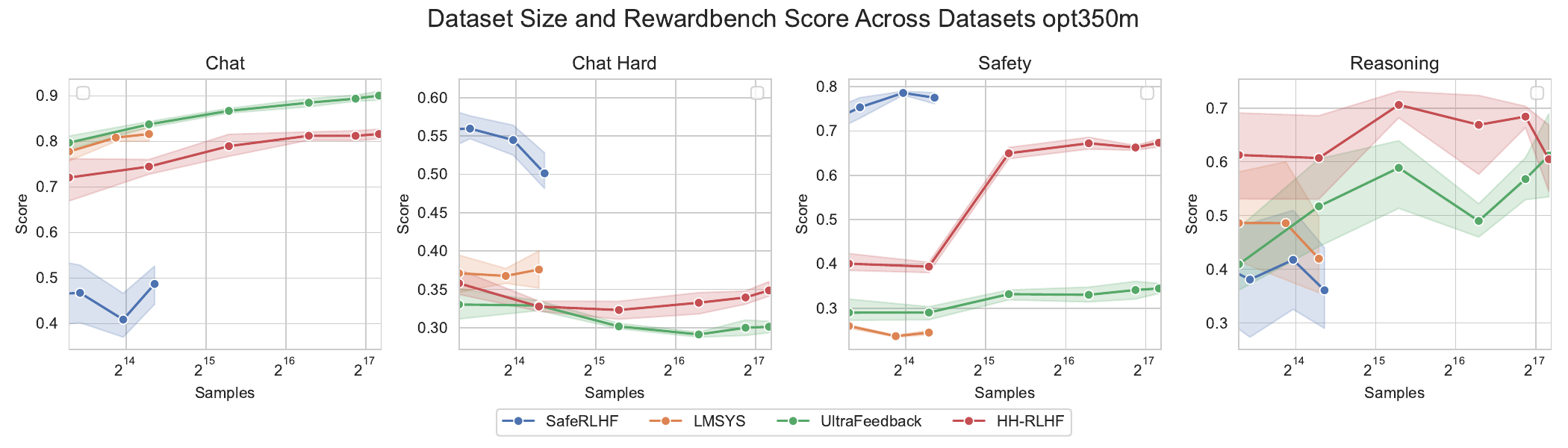}
    \caption{Comparing RewardBench performance across different datasets for OPT350M Model. Increasing the dataset size is insufficient to close the performance gap between datasets the best dataset depends on the evaluation task within RewardBench.}
    \label{fig:ood-scaling-350m}
\end{figure}

In the main text we show the OOD performance for the Llama2-7B-Chat model. We also include the Llama2-7B base model (Figure \ref{fig:ood-scaling-7B}) where we see the same pattern of \textsc{UltraFeedback} dominating the chat category and \textsc{SafeRLHF} dominating the safety category of Rewardbench. For smaller models, Figure~\ref{fig:ood-scaling-1B}, shows a similar pattern for TinyLlama-1B and Figure~\ref{fig:ood-scaling-350m}. In these smaller models, the advantage of the \textsc{SafeRLHF} is even more stark. 

\subsection{Noise Invariance}
\label{app:noise}
We also include plots of the effect of dataset label noise on Rewardbench tasks. For all of the models, the Chat and Safety tasks are not significantly affected until 40\% of the labels are flipped. For the Chat Hard and reasoning tasks, most of the models we train are not good enough to examine differences properly. It is also interesting that we do not observe cross-over behavior; no dataset starts with a worse performance and improves over a different dataset at a higher level of noise. 
\begin{figure}
    \centering
    \includegraphics[width=\linewidth]{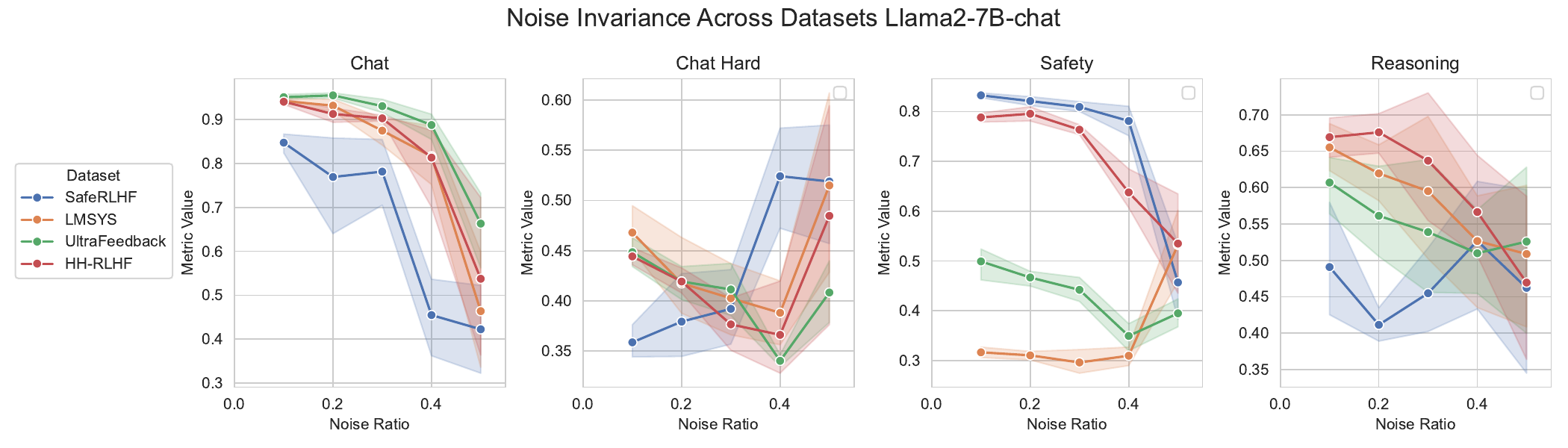}
    \caption{Comparing RewardBench performance across different datasets for Llama2-7B-chat reward model for different levels of label noise. Performance is relatively stable until 30\% of labels have been flipped.}
    \label{fig:noise-rb-llama-7b-chat}
\end{figure}

\begin{figure}
    \centering
    \includegraphics[width=\linewidth]{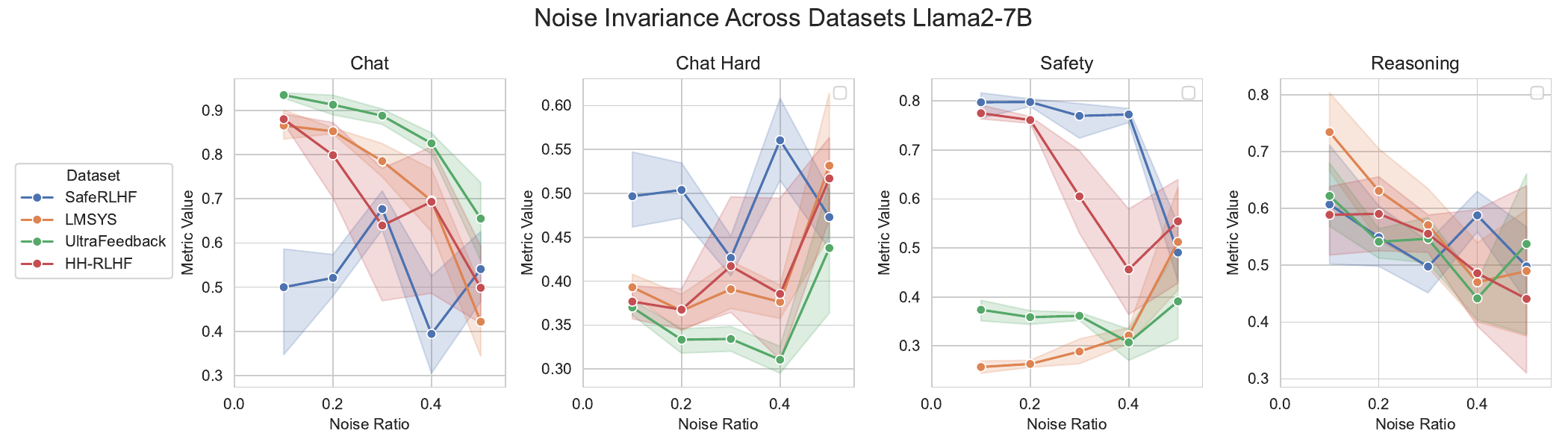}
    \caption{Comparing RewardBench performance across different datasets for Llama2-7B reward model for different levels of label noise. Performance is relatively stable until 30\% of labels have been flipped.}
    \label{fig:noise-rb-llama-7b}
\end{figure}

\begin{figure}
    \centering
    \includegraphics[width=\linewidth]{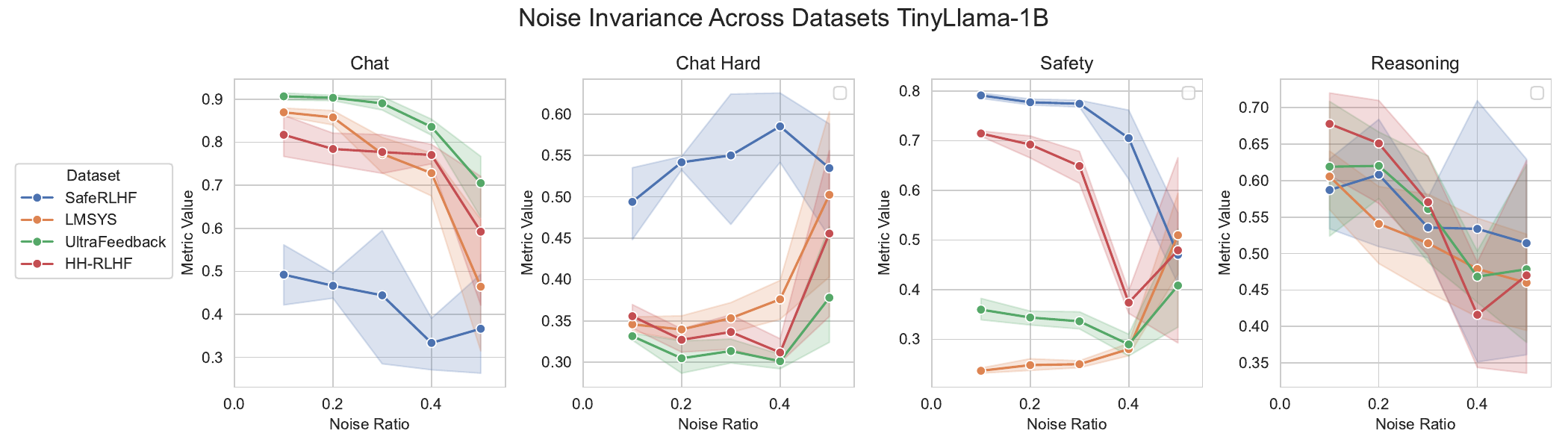}
    \caption{Comparing RewardBench performance across different datasets for TinyLlama-1B reward model for different levels of label noise. Performance is relatively stable until 30\% of labels have been flipped.}
    \label{fig:noise-rb-llama-1b}
\end{figure}

\begin{figure}
    \centering
    \includegraphics[width=\linewidth]{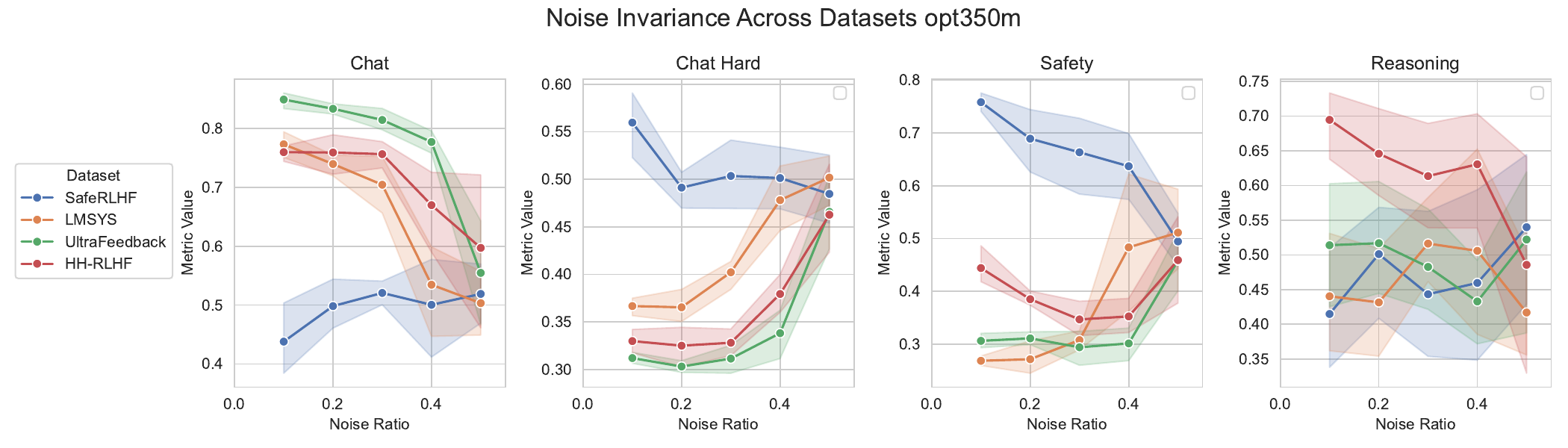}
    \caption{Comparing RewardBench performance across different datasets for Opt-350m reward model for different levels of label noise. Performance is relatively stable until 30\% of labels have been flipped.}
    \label{fig:noise-rb-opt350m}
\end{figure}